\documentclass[sigconf]{acmart}
\usepackage{float}
\usepackage[belowskip=-1pt,aboveskip=0pt]{caption}
\def\BibTeX{{\rm B\kern-.05em{\sc i\kern-.025em b}\kern-.08emT\kern-.1667em\lower.7ex\hbox{E}\kern-.125emX}}
    
\copyrightyear{2019}
\acmYear{2019}
\setcopyright{acmcopyright}
\acmConference[GeoAI'19]{3rd ACM SIGSPATIAL International Workshop on AI for Geographic Knowledge Discovery }{November 5, 2019}{Chicago, IL, USA}
\acmBooktitle{3rd ACM SIGSPATIAL International Workshop on AI for Geographic Knowledge Discovery (GeoAI'19), November 5, 2019, Chicago, IL, USA}
\acmPrice{15.00}
\acmDOI{10.1145/3356471.3365238}
\acmISBN{978-1-4503-6957-2/19/11}

 \renewcommand\footnotetextcopyrightpermission[1]{} % removes footnote with conference information in first column
\begin{document}

%
% The "title" command has an optional parameter, allowing the author to define a "short title" to be used in page headers.
%\title{An unsupervised approach to Geographical Knowledge Discovery using street level images and street network images}
\title[An unsupervised approach to geographic knowledge discovery]{An unsupervised approach to geographical knowledge discovery using street level and street network images}

\author{Stephen Law$^{*}$}
\affiliation{%
 \institution{The Alan Turing Institute  \& \linebreak University College London}
 \city{London}
 \country{UK}}
  
\author{Mateo Neira$^{*}$}
\affiliation{%
 \institution{The Alan Turing Institute \& \linebreak University College London}
 \city{London}
 \country{UK}}

% comments 
%* details of the latent variable regression and autoencoder in appendix
%* streets - more samples needed. tuning of mlp. regression with degree, closeness, betweenness, intelligibility, synergy

% By default, the full list of authors will be used in the page headers. Often, this list is too long, and will overlap
% other information printed in the page headers. This command allows the author to define a more concise list
% of authors' names for this purpose.

\renewcommand{\shortauthors}{Law and Neira}

%
% The abstract is a short summary of the work to be presented in the article.
\begin{abstract}

Recent researches have shown the increasing use of machine learning methods in geography and urban analytics, primarily to extract features and patterns from spatial and temporal data using a supervised approach. Researches integrating geographical processes in machine learning models and the use of unsupervised approaches on geographical data for knowledge discovery had been sparse. This research contributes to the ladder, where we show how latent variables learned from unsupervised learning methods on urban images can be used for geographic knowledge discovery. In particular, we propose a simple approach called Convolutional-PCA ($ConvPCA$) which are applied on both street level and street network images to find a set of uncorrelated and ordered visual latent components. The approach allows for meaningful explanations using a combination of geographical and generative visualisations to explore the latent space, and to show how the learned representation can be used to predict urban characteristics such as street quality and street network attributes. The research also finds that the visual components from the $ConvPCA$ model achieves similar accuracy when compared to less interpretable dimension reduction techniques.

\end{abstract}

 \begin{CCSXML}
<ccs2012>
<concept>
<concept_id>10010147.10010257.10010258.10010260</concept_id>
<concept_desc>Computing methodologies~Unsupervised learning</concept_desc>
<concept_significance>500</concept_significance>
</concept>
<concept>
<concept_id>10010147.10010178.10010224</concept_id>
<concept_desc>Computing methodologies~Computer vision</concept_desc>
<concept_significance>300</concept_significance>
</concept>
<concept>
<concept_id>10010405.10010469.10010472</concept_id>
<concept_desc>Applied computing~Architecture (buildings)</concept_desc>
<concept_significance>500</concept_significance>
</concept>
</ccs2012>
\end{CCSXML}

\ccsdesc[500]{Computing methodologies~Unsupervised learning}
\ccsdesc[300]{Computing methodologies~Computer vision}
\ccsdesc[500]{Applied computing~Architecture (buildings)}

%
% Keywords. The author(s) should pick words that accurately describe the work being
% presented. Separate the keywords with commas.
\keywords{urban analytics, unsupervised learning, convolutional neural networks, knowledge discovery, computer vision, machine learning}
\thanks{$^{*}$Both authors contributed equally to this research.}

%
% This command processes the author and affiliation and title information and builds
% the first part of the formatted document.
\maketitle

\section{Introduction}
According to \cite{Miller:2001:GDM:940614}, Geographic knowledge discovery (GKD) is the process of using computational methods and visualisation to explore spatial databases to discover useful geographic knowledge. Despite the ubiquity of geographically-labelled image data and the subsequent use of machine learning methods to retrieve geographical information, the majority of the researches have focused mainly on the use of supervised learning approaches. For example, on the use of convolutional neural networks $CNN$ to make inferences on perceived safety \cite{Naik2013}, house price \cite{Law2018} and scenicness \cite{Seresinhe2017}. These researches required effort on both collecting the data and on learning a specific objective. As a result, there is an opportunity to use urban image information in an unsupervised and scalable manner. Our research question therefore implies, what compact latent representation can be learnt from urban images without supervision, how can this information be described and what is this representation useful for? 

%With the exception of some recent efforts, there had been a lack of research on integrating geographical theory and geographical data to better understand machine learning methods. 

This study contributes to these research questions and proposes an unsupervised learning model called Convolutional-PCA ($ConvPCA$) that summarises urban imagery into a set of lower dimensional uncorrelated latent components. We apply this method to two case studies namely for: Google StreetView images\cite{GoogleStreetView} and OpenStreetsMaps ($OSM$) street network images \cite{OpenStreetMap}. In the experiments, we first map and visualise the extremes of the responses geographically and generate new synthetic images by perturbing the values of each component whilst holding all the other component values constant. We then study the latent components by using it to predict different geographical datasets such as street enclosure and street frontage type for the StreetView image data and network density and network centrality for the street network image data. The research finds that the visual components from the $ConvPCA$ model have interpretable meanings with predictive abilities to geographical labelled data using a compact representation. The research also finds that the visual components from the $ConvPCA$ model achieve a similar accuracy to other dimension reduction techniques such as an autoencoder while retaining its interpretability. From a machine learning perspective, we gain new knowledge about these latent components which contribute to the recent efforts in linking the two disciplines \cite{Klemmer2019} \cite{Reichstein2018}. 

\section{Related Works}

\subsection{StreetViews}
Street-level images have been used extensively in intelligent transportation systems research. Specifically on the deployment of autonomous vehicles where deep convolutional neural networks ($CNN$) had been applied for urban scene understanding \cite{Ros2016}. More recently, we have also seen the use of generative models such as Generative Adversarial Networks ($GAN$) to synthetically create street scenes that can be used to train self-driving vehicles \cite{vid2vid}. Despite its popularity in transportation research, there had been limited effort on using street-level imagery to retrieve geographical information and for studying urban planning problems. One such example is StreetScore where \cite{Naik2013} collected subjective human perception data from street images through a crowd-sourced survey (Place Pulse 2.0) which are then used to predict the perceived safety of a place \cite{Dubey2016}. Another example is the work of Gebru et al. \cite{Gebru2017} whom extracted features such as car types from Google StreetView images to predict the income, race, education, and voting patterns for cities in the United States. We have also seen the use of urban images \cite{Seresinhe2017} to predict scenicness ratings which were found to affect urban wellbeing as well as the type of urban frontages \cite{Law2018} which is an important urban design attribute. These fairly recent efforts relied on extracting visual features from street-level images which are then related to different socio-economic factors. In contrast to these works, Law et al \cite{Law2019} extracted a visual response from urban images by directly estimating house prices. A distinguishing difference here is that the method extracted a visual scalar response that corresponds directly to house price, which can be visualised and interpreted in traditional econometric model. In summary, these recent researches focused on learning a set of visual features or response from urban imagery using a supervised learning approach. Our research extends from this work where we propose a two stage method in learning a set of generic and compact latent visual latent components from an unsupervised learning approach. We then, through a set of analysis and experiments interrogate, describe and explore these components for geographic knowledge discovery. 

\subsection{Street Networks}
In the case of street networks, there has been a long-standing effort to analyse and to understand them from a quantitative perspective and to generate models that are able to reproduce their empirical features. Previous works have largely been based on complexity theory and network science \cite{Boeing2018, louf2014typology, strano2012elementary}. This includes analyzing the spatial configuration of urban street networks \cite{hillier2007space} and analyzing urban systems from an information theoretic perspective \cite{batty2005}.  

More recently, there has been a growing interest in applying machine learning methods to extract useful information from the vast amount of data now openly available from sources such as $OSM$. Examples of such works have used neural networks to classify street network patterns of different cities, where two different methods had been used. The first used a Convolutional Autoencoder $CAE$ to create dense urban vectors that are used to cluster similar urban morphologies using a self-organinzing map \cite{Moosavi2017}. The second approach used a Variational Auto Encoder $VAE$ to measure similarity across different networks \cite{Kempinska2019}. 

Generative models have also been used to generate synthetic street networks. Variational Autoencoder trained on street network images has been use by sampling from the latent space $z$ \cite{Kempinska2019}, however the resolution of these are low, and fail to capture fine grain detail of local streets. A Generative Adversarial Network such as $StreetGAN$ \cite{Hartmann2017} has also been proposed to generate a multitude of arbitrary sized street networks that faithfully reproduce the style of the original dataset. 

Current limitations in the use of VAE, CAE, and GANs on street networks lie in the interpretability of the latent space and its relationship to geometrical and topological properties used in established network measures. Our research contributes to these researches by developing a methodology to interpret the lower-dimensional embedding learnt by a convolution autoencoder. This allows for greater interpretation of the unsupervised model, as well as providing some initial results as to the relationship between the embedding and the established network measures. 

\section{Methods and Materials}

\subsection{Convolutional-PCA}
\begin{figure}[!ht]
	\centering 
	\includegraphics[width=1\linewidth] {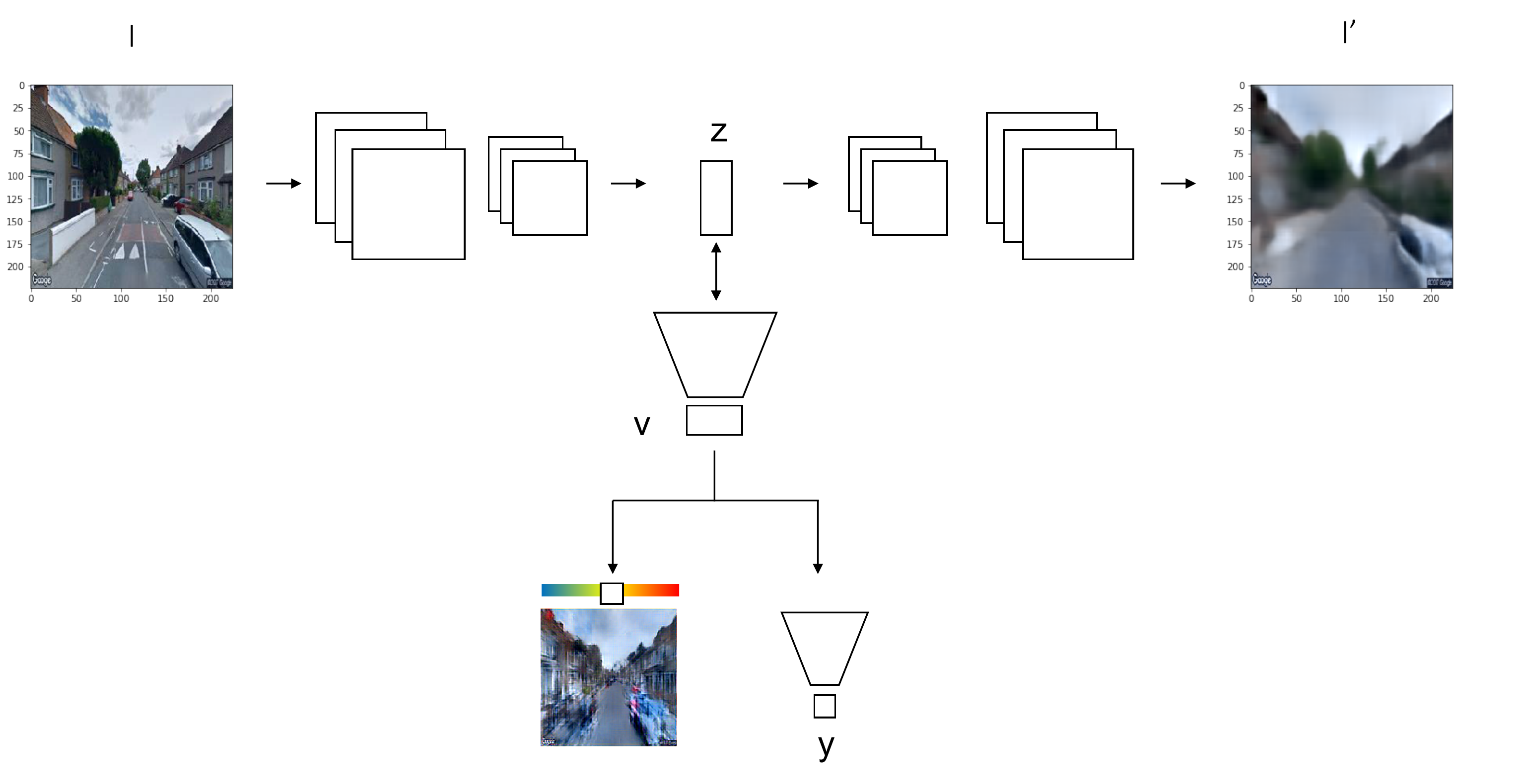}
	\caption{Architecture of ConvPCA, which combines a Convolutional AutoEncoder ($CAE$) with a linear PCA ($PCA_{lin}$) to retrieve a set of uncorrelated and ordered visual latent components that summarises street-level and street-network image data. These latent components are then used for knowledge discovery through visualisation and different down-stream tasks such as prediction and classification.}
	\label{fig:Arch}
\end{figure}

We propose here the Convolutional-PCA ($ConvPCA$), which combines a type of Convolutional Neural Network called the Convolutional Autoencoder ($CAE$) with a linear PCA ($PCA_{lin}$) to retrieve a set of latent visual components that summarise a StreetView image or a street networks image. We first describe the $CAE$ followed by the $PCA_{lin}$. Deep Convolutional Autoencoder $CAE$ is an unsupervised method that uses convolutional neural network ($CNN$) to learn a compact representation or a set of visual features \cite{CAE, AE, Guo2017}. Deep CAE consists of two set of layers, an encoder $f_w( \cdot )$ and a decoder $g_u( \cdot )$

\begin{equation}
  \begin{array}{l}
    f_w(x)=\sigma (x\ast W) \equiv z \\
    \\
    g_u(z)=\sigma (z\ast U)
  \end{array}
\end{equation}

where $x$ is the input vector, $z$ is the latent features, $\ast$ is the convolution operator that extract image features and $\sigma$ is a $ReLU$ activation function to model nonlinearity in the neural network. These convolutional layers can be stacked sequentially where the encoding layers reduce the dimension to a latent variable $z$ while the decoding layer increases the dimension back to image space. The sequential architecture can be seen in figure \ref{fig:Arch}. Following \cite{CAE}, the parameters of the encoder $z=F_w(x)$ and the decoder $x^\prime = G_{u}$ are updated by minimising the reconstruction losses between $x$ and $x^\prime=G_{u}(F_w(x_i))$.

\begin{equation}\label{eq:Loss}
L_r= \frac{1}{n} \sum (x_i-G_{u}(F_w(x_i)))^2
\end{equation}

In our research, we further compress the latent visual features by applying a linear principal component analysis $PCA_{lin}$ which summarises the visual feature $z$ into a set of linearly uncorrelated variables $v$. To compute $v$, we first standardise $z$ and compute the eigenvectors and eigenvalues of the feature covariance matrix $P$. We then take the eigenvectors to calculate the full principal component decomposition of $z$, given by $V=XW$, where $W$ is the eigenvector matrix. $V$ can be re-projected back on to the original latent space produced by the encoder before passing in to the decoder to reconstruct the images. This process allows us to:

\begin{itemize}
	\item Retrieve a set of uncorrelated and ordered visual latent components that can be visualised and mapped geographically.
	\item Make changes to individual components and decoding it to generate a synthetic image. %to test their response in the decoder.
	\item Relate learnt visual latent components to geographical labelled data.
\end{itemize}

To discover new geographical knowledge and in testing the usefulness of the latent representation, we will visualise these components by generating new images when perturbing in the $PCA_{lin}$ space and also in mapping them. We will then use these components for down stream tasks such as prediction and classification. The process can be seen in figure \ref{fig:Arch}. Further research is required to validate the meaning of these visual latent components quantitatively and in comparing it with latent components extracted from other methods. These limitations will be further elaborated in the discussion section. 

$PCA_{lin}$ is selected as it is a well principled dimension reduction technique that learns a compact and meaningful representation with uncorrelated and ordered components. Approaches such as autoencoders can find a similar representation but without the same interpretability \cite{Ladjal}. In the prediction experiment, we will study and compare the extent a $PCA_{lin}$, a linear autoencoder and a non-linear autoencoder are able to learn a compact representation for different down-stream tasks. A benefit of finding uncorrelated ordered components is that these factors can be inserted into a generalised linear modelling framework, whose coefficients can be interpreted. Such research are not explored in this paper but the coefficients are meaningful for example in econometric studies \cite{Law2019}.

\subsection{Materials}
\begin{figure}[!ht]
	\centering 
	\includegraphics[width=1\linewidth] {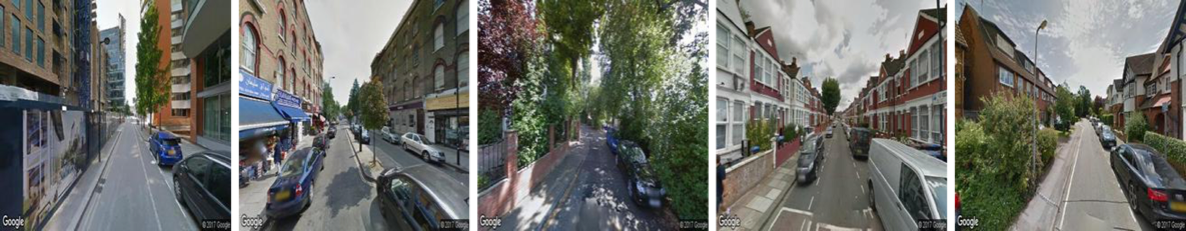}
	\caption{Examples of street level images from Google StreetView. \copyright 2017 Google Inc.}
	\label{fig:streetviews_sample}
\end{figure}

We collected two datasets. The first dataset is street images taken from the Google StreetView API~\cite{GoogleStreetView}\footnote{\copyright 2017 Google Inc.}. Similar to \cite{Law2018}, we collected a front-facing image for each street in the Greater London Area. To collect the dataset, we constructed a line-graph from the street network of London (OS Meridian line2 dataset \cite{OS}). We then take the geographic median and the azimuth of the street edge to give both the location and the bearing when collecting each image. We collected a total of $110,493$ street images in London. For more details in the data collection method please see \cite{Law2018}. Figure \ref{fig:streetviews_sample} illustrates typical images from the dataset. 

\begin{figure}[!ht]
	\centering 
	\includegraphics[width=1\linewidth] {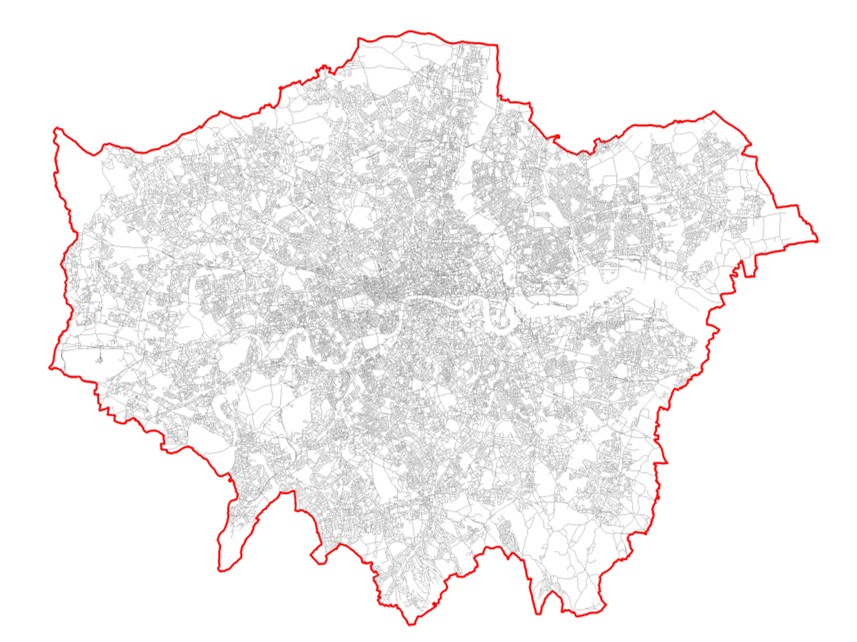}
	\caption{The Greater London case study area boundary.}
	\label{fig:GLA_boundary}
\end{figure}

\begin{figure}[!ht]
	\centering 
	\includegraphics[width=1\linewidth] {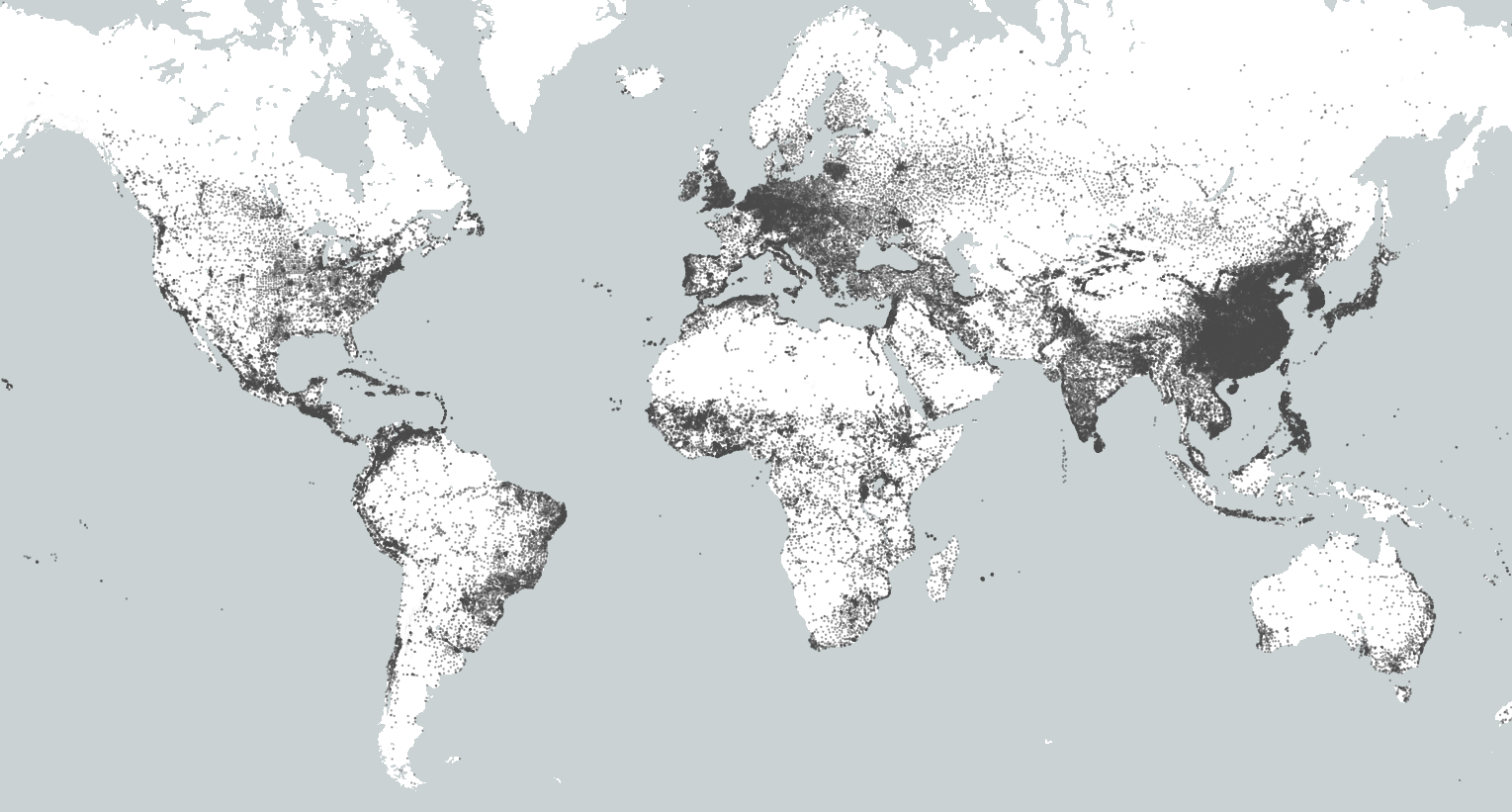}
	\caption{Centroid of 107,973 cities and towns used for training and testing.}
	\label{fig:cities}
\end{figure}

\begin{figure}[!ht]
	\centering 
	\includegraphics[width=1\linewidth] {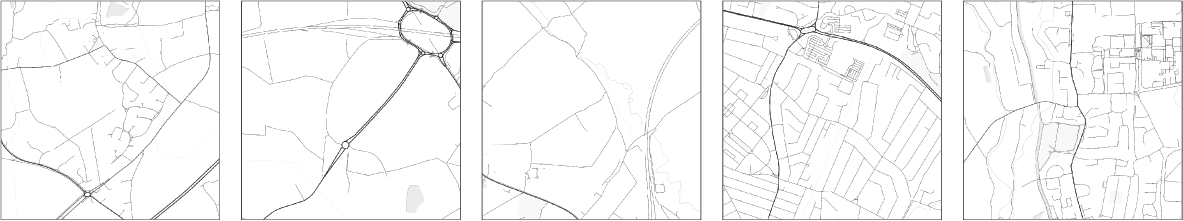}
	\caption{Sample of rasterized street network data.}
	\label{fig:streets_sample}
\end{figure}

The second dataset is the street network dataset taken from OpenStreetsMaps \cite{OpenStreetMap}, we query all the cities and towns for a total of $107,973$. For each city and town we download the street network within a 1.5km x 1.5Km box at the centroid of each place using osmnx \cite{boeing2017osmnx}, as shown in Figure \ref{fig:cities}. For each 1.5km x 1.5km grid we retrieve a graph $G=(V,E)$ where each vertex $v$ corresponds to a street intersection and $e$ edge corresponds to a street segment. For each $G$, we rasterise it into a 256 x 256 pixel image as shown in Figure \ref{fig:streets_sample}. We also calculate basic network statistics \cite{boeing2018multi} such as network centrality and network density that are later used to test the learnt features of the images.

\section{Experimental Results}
In order to discover new information and interpretations from these visual latent components, we will visualise these components and to use these factors for predictions on both the street level and street network dataset.

\subsection{streetview images}
\subsubsection{Visualisation experiments}

The $ConvPCA$ first learns a mapping from a three channel street level images (224 x 224 x 3) down to a lower dimensional embedding (4,096 dimensions) using a convolutional autoencoder $CAE$. The lower dimension embedding is then further summarised into a set of uncorrelated components using $PCA_{lin}$. For the StreetView images, we adopted a $VGG$ \cite{VGGnet} like architecture where we keep the kernel size and filter numbers constant across both the encoder and decoder. 

To show the results, we first plot the images with the highest and lowest principal component values for interpretation. In this case, component $pca$ 7 has blank facade in one of the extremes and natural scenes in the other. $pca$ 10 and $pca$ 30 shows a tunnel space in one extreme and a mixture of urban scenes in the other. While $pca$ 14 has buildings in one extreme and blank facades in the other. Lower rank components that capture lower variance seem to be showing less patterns and therefore not visualised. 

\begin{figure}[!ht]
	\centering 
	\includegraphics[width=1\linewidth] {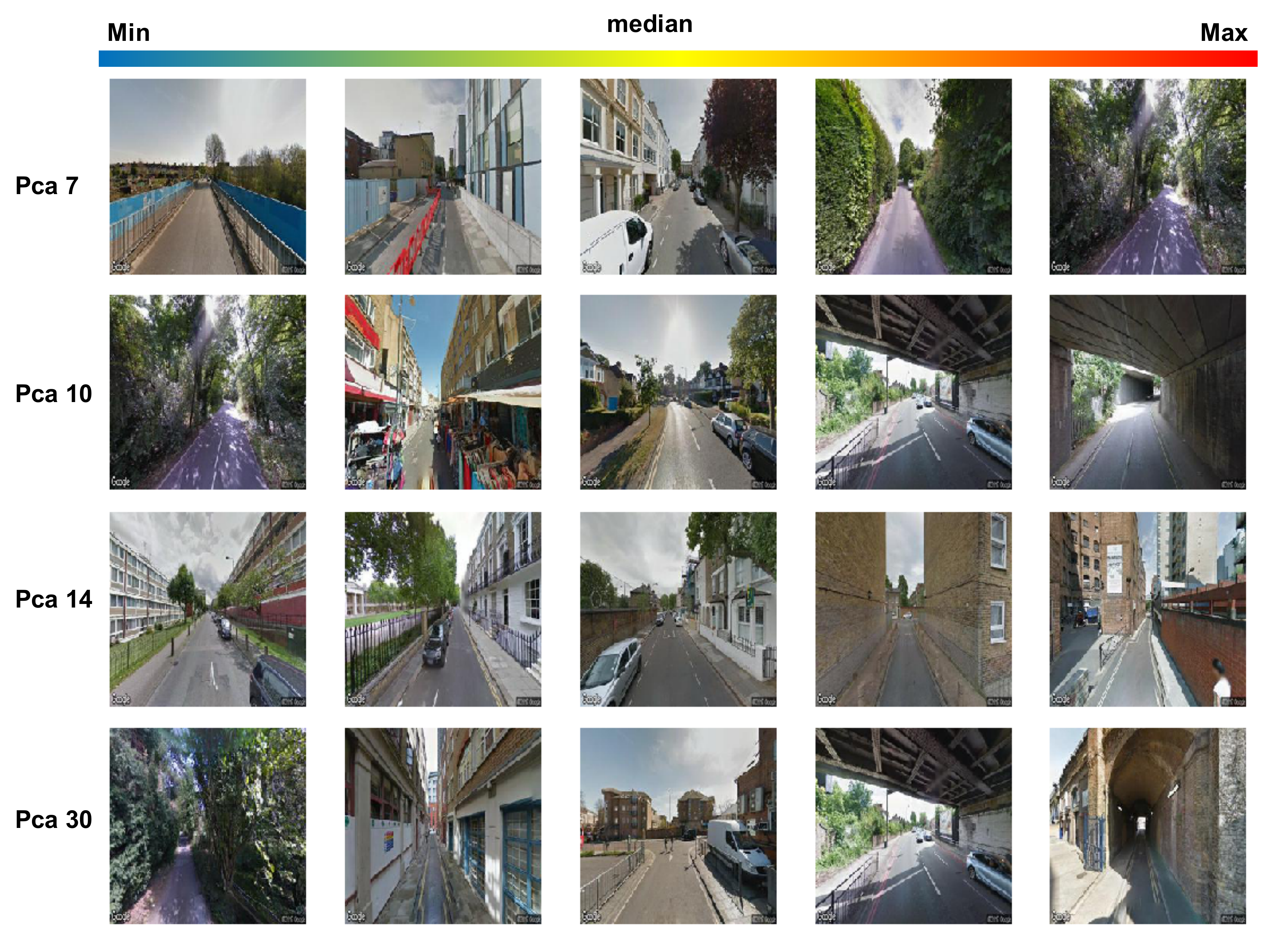}
	\caption{Example London Google StreetView images according to its component values. $pca$ 7  has blank facade in one of the extremes and natural scenaries in the other. $pca$ 10 and $pca$ 30 shows a tunnel space in one extreme and a mixture of urban scenaries in the other. $pca$ 14 shows buildings in one extreme and blank facades in the other. \copyright 2017 Google Inc.}
	\label{fig:vis01}
\end{figure}

To interrogate the results of the primary components, we focus on visualising $pca$ 1 and $pca$ 3 geographically in figure \ref{fig:vis01}. The images plotted above the map show the two extremes of the visual latent components. We can descriptively interpret these two visual components $v_1$ and $v_3$ as proxy measures for different type of street urbanity. We also show through global spatial autocorrelation analysis these components exhibit strong spatial dependence. Please see the $appendix$ for more details on the spatial analysis.

\begin{figure}[!ht]
	\centering 
	\includegraphics[width=1\linewidth] {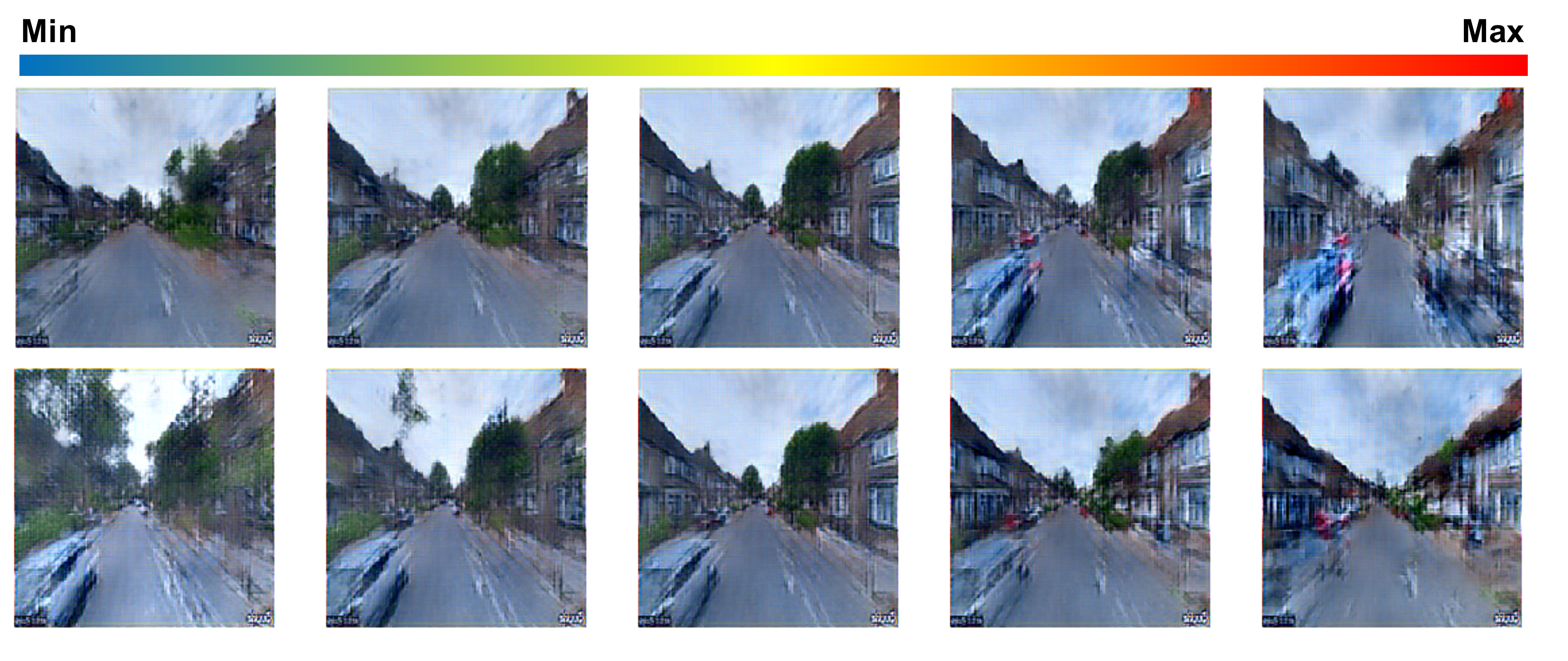}
	\caption{Visual latent component perturbations of a typical StreetView image. The first row shows the perturbation of $pca$ 1 where building details increase in one end of the axis and reduces on the other. The second row shows the perturbation of $pca$ 3 where greenery increase on one end of the axis and building density on the other end of the axis. \copyright 2017 Google Inc.}
	\label{fig:streetview_changes}
\end{figure}

We then visualised one of the StreetView images and perturbed each of the two principal components while holding all the other component values constant before passing it to the decoder to generate a synthetic image. More formally, for each $pca$ we create a mean vector $\hat{v}$, where we keep all values in $\hat{v}$ constant and vary only the individual $pca$ before passing it to the decoder to generate a synthetic image. Figure \ref{fig:streetview_changes} shows when we perturbed $pca$ 1 of a typical StreetView image while holding all other principal components constant, building details tends to increase, and when we perturbed the same image in the other axis, building details tend to reduced. In contrast, when we perturbed $pca$ 3 of the same StreetView image whilst holding all other principal components constant, trees started appearing and when we perturbed the same image to the other axis buildings becomes more prominent and the trees disappeared. The result also shows that the streets are widening in one of the axis while the car is disappearing in the other axis for $pca$ 1. This result suggests, each component is related to a quality measure of street urbanity and is possibly capturing multiple correlated visual features of a StreetView image. As a result, in terms of controllability, the approach seems not able to disentangle highly correlated features. These descriptive results show geographical and generative visualisations are useful approaches to discover meanings from these visual latent components. However, more researches is needed to validate the meaning of these visual components quantitatively. %, improving the quality of the image reconstruction using a generative model and to provide more control in disentangling highly correlated visual features. 

\begin{figure}[!ht]
	\centering 
	\includegraphics[width=1\linewidth] {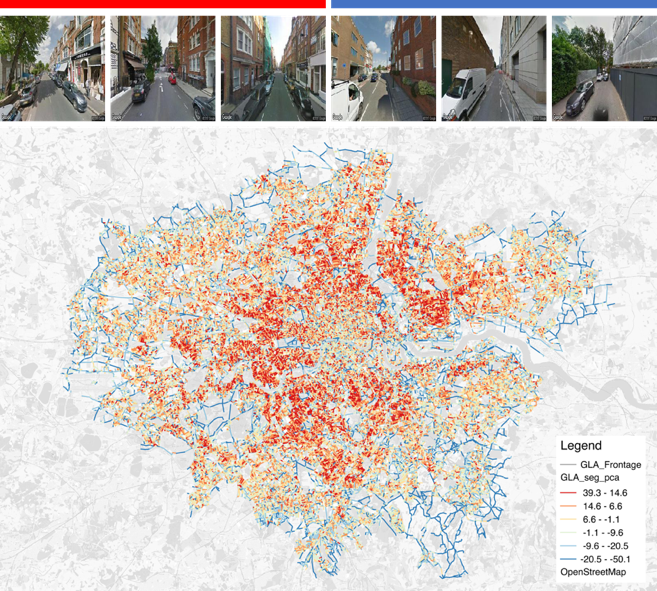}
	\caption{$pca$ 1 shows descriptively a quality measure for street urbanity. The map shows that the intra-urban area has higher component value. Visualising the extremes of the component shows greater building details in one end of the axis.\copyright 2017 Google Inc.}
	\label{fig:vis03}
\end{figure}

\begin{figure}[!ht]
	\centering 
	\includegraphics[width=1\linewidth] {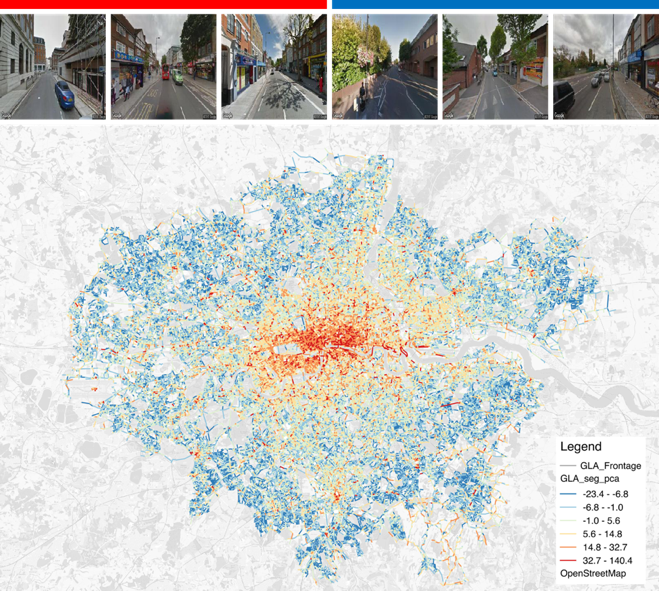}
	\caption{$pca$ 3 shows descriptively another quality measure for street urbanity. The map shows central London has higher component values and outer London has less. Visualising the extremes of the latent component shows higher building density in one end of the axis.\copyright 2017 Google Inc.}
	\label{fig:vis04}
\end{figure}

%\begin{figure*}[!h]
%	\centering
%  \begin{tabular}{p{\columnwidth}p{\columnwidth}}
%	  \includegraphics[width=1\columnwidth] {figs/GLA_Vis01.png}
%    &
%	  \includegraphics[width=1\columnwidth] {figs/GLA_Vis02.png}
%  \end{tabular}
%	\caption{{\bf Left:} London StreetView Ornateness. We interpret this component as a measure of building ornateness where red denotes higher visual richness and blue denotes lower visual richness. {\bf Right:} London Urban Density. We interpret this component as a measure of urban density or building intensity where red denotes higher density and blue denotes lower density.
%	\label{GLA}	\label{fig:GLA_Vis}
% Contains Ordinance Survey data \copyright Crown copyright and database right \copyright 2017.}
%\end{figure*}

%\begin{figure}[!ht]
%	\centering 
%	\includegraphics[width=1\linewidth] {figs/street_changes.png}
%	\caption{Latent visual component perturbations of a London StreetView image. By perturbing the first principal component of a typical StreetView, we show greater building details in one of the extremes and greater urban greenery in the other extreme.}
%	\label{fig:streetview_changes}
%\end{figure}

\subsubsection{Prediction experiment}

In order to demonstrate the usefulness of these visual latent components for different down-stream tasks, we constructed two separate models to map the latent visual components $V$ to street enclosure (regression task) and street frontage quality (classification task). We compare $PCA_{lin}$ to two other dimension reduction techniques in retrieving the latent visual component, a linear autoencoder $AE_{lin}$ and a nonlinear autoencoder $AE_{non}$. The linear autoencoder uses a linear activation function with one bottleneck layer that outputs $V$. The nonlinear autoencoder on the other hand uses the $ReLu$ activation function with three hidden layers where the first and the third hidden layer are the encoding and decoding layer with $512$ neurons and the second layer being the bottleneck layer that outputs $V$. 

\begin{figure}[!ht]
	\centering 
	\includegraphics[width=1\linewidth] {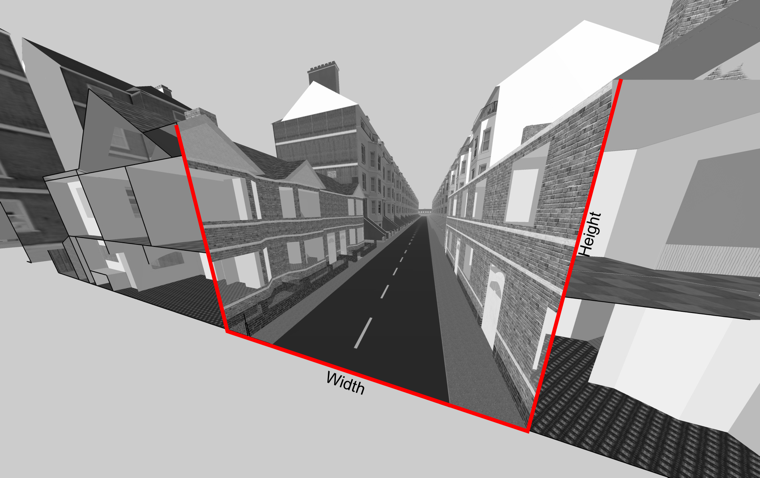}
	\caption{Street enclosure diagram. We define street enclosure as the ratio between $avg.height/avg.width$. }
	\label{fig:enclosure}
\end{figure}

Street enclosure here is defined as the average height of the building of a street divided by the average width between the buildings of the same street as illustrated in fig \ref{fig:enclosure}. The street enclosure is calculated by segmenting the streets from Ordnance Survey data \cite{OS} every $40m$. For each street segment $S$, we calculate the geographic median $S_{c}$ and the azimuth $S_{\alpha}$, and create a new line that is perpendicular to $S$ at the point $S_c$. The perpendicular line $S_{\bot}$ is used to create the street profile by intersecting it with the closest building on either side of the street and querying the associated height attribute, this is used to calculate the street enclosure as building height to street width ratio $enc=\overline{h}/w$. Please see \cite{Neira2019} for additional details.

Street frontage types here is defined with four frontage categories namely, active frontage on both sides of the street, active frontage on one side of the street, non-active frontage and non urban frontage. This dataset was manually compiled and studied from a previous study. Please see \cite{Law2018} for additional details.

We split the dataset randomly into a train (70\%), validation (15\%) and test set (15\%). We then train a multi-layer perceptron $F(\cdot)$ to predict street enclosure and street frontage types from the visual latent components $V$ as inputs , parameterized by a set of weights $W_{v}$. 

The multi-layer perceptron ($mlp$) here is defined as a fully connected neural network with three hidden layers. The first fully connected layer has 64 hidden nodes, while the second layer has 32 hidden nodes and the third layer has 16 hidden nodes. A dropout layer ($0.2$) and a $l1$ regularisation was added in the final activation layer for better generalisation. To test the importance of the visual components with respect to the model accuracy, we constructed five different models based on the number of components [4,8,16,32,64] using the three dimension reduction techniques. This results in 15 models in total. 

We train the street enclosure model to minimize the mean squared error $mse$ on a training set, using the ADAM \cite{ADAM2014} optimizer with an initial learning rate set at 0.001. We then report the mean squared error ($MSE$) and the coefficient of determination $R^2$ between the model prediction and the observed street enclosure for the spatially random test-set. Similarly, we train the street frontage model to minimise the categorical cross-entropy losses on the training set, using ADAM \cite{ADAM2014} optimiser with an initial learning rate set at 0.001. We then report the cross entropy losses and the accuracy which is simply the sum of correctly predicted frontage class over all samples. All the experiments are conducted with the Keras library~\citep{chollet2015} using a Tensorflow~\cite{tensorflow2015-whitepaper} back-end.

\begin{figure}[!ht]
	\centering 
	\includegraphics[width=1\linewidth] {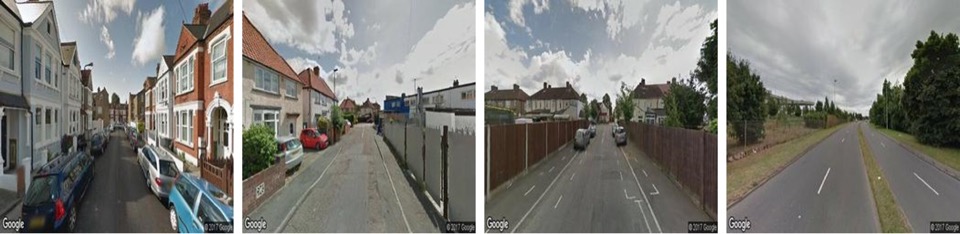}
	\caption{Street frontage diagram. We classify street images into four street frontage categories namely $both-side-active$, $single-side-active$, $non-active$, $non-urban$ \cite{Law2018}.\copyright 2017 Google Inc.}
	\label{fig:frontage}
\end{figure}

The results in table~\ref{tab:run1} shows the $losses$ and $accuracy$ of the three dimension reduction techniques when predicting street enclosure for a spatially random test-set. The model with 64 components achieve 60\% accuracy, while the model with 4 components achieve 40-50\% accuracy. The result shows, we can achieve similar levels of accuracy with $PCA_{lin}$ when compared to both $AE_{lin}$ and $AE_{non}$.  Similarly, the results in table~\ref{tab:run2} shows the $losses$ and $accuracy$ of the three dimension reduction techniques when predicting street frontage quality for a spatially random test-set. The results show a model with more components achieve a higher accuracy ($70\%$) than one with less and that $PCA_{lin}$ achieves comparable accuracy to $AE_{lin}$ and $AE_{non}$. These results suggest, the convolutional layers are possibly capturing some of the non-linear effects between the different image features in the data. As a result, a linear dimension reduction technique such as $PCA_{lin}$, is able to learn a compact representation of the latent variable $z$ which captures similar variance for two predictive tasks when compared to the  autoencoders while retaining its interpretability.

\begin{table}[!ht]
\centering
\caption{Street Enclosure Results \label{tab:run1}}

\begin{tabular}{r|c|c|c}
%\toprule
%\hline
\textbf{}  $accuracy$      & $PCA_{lin}$      & $AE_{lin}$ & $AE_{non}$     \\ \hline
4 components    & 41.50\%   &   41.80\%         & 50.96\%         \\ \hline
8 components    & 53.64\%   &   54.02\%         & 55.78\%          \\ \hline
16 components   & 56.37\%   &   56.24\%         & 56.77\%         \\ \hline
32 components   & 58.36\%   &   58.00\%         & 58.62\%          \\ \hline
64 components   & 59.17\%   &   58.45\%         & 59.98\%          \\ \hline
           \\ 
%\hline
\end{tabular} \vspace{0.5em}

\begin{tabular}{r|c|c|c}
%\toprule
%\hline
\textbf{}  $losses$    & $PCA_{lin}$    & $AE_{lin}$ & $AE_{non}$     \\ \hline
4 components    & 0.603   &   0.600         & 0.499         \\ \hline
8 components    & 0.480   &   0.474         & 0.450          \\ \hline
16 components   & 0.450   &   0.451         & 0.440         \\ \hline
32 components   & 0.429   &   0.433         & 0.421          \\ \hline
64 components   & 0.421   &   0.428         & 0.407          \\ \hline
           \\ 
%\hline
\end{tabular} \vspace{0.5em}
\end{table}

\begin{table}[!ht]
\centering
\caption{Street Frontage Results \label{tab:run2}}

\begin{tabular}{r|c|c|c}
%\toprule
%\hline
\textbf{}  $accuracy$      & $PCA_{lin}$      & $AE_{lin}$ & $AE_{non}$     \\ \hline
4 components    & 61.46\%   &   59.91\%         & 60.96\%         \\ \hline
8 components    & 64.10\%   &   62.26\%         & 62.84\%          \\ \hline
16 components   & 68.24\%   &   67.17\%         & 67.44\%         \\ \hline
32 components   & 69.13\%   &   68.51\%         & 68.71\%          \\ \hline
64 components   & 71.41\%   &   71.93\%         & 70.50\%          \\ \hline
           \\ 
%\hline
\end{tabular} \vspace{0.5em}

\begin{tabular}{r|c|c|c}
%\toprule
%\hline
\textbf{}  $losses$    & $PCA_{lin}$    & $AE_{lin}$ & $AE_{non}$     \\ \hline
4 components    & 0.884  &   0.898         & 0.907         \\ \hline
8 components    & 0.839   &   0.846        & 0.858          \\ \hline
16 components   & 0.758   &   0.759        & 0.772         \\ \hline
32 components   & 0.734   &   0.735         & 0.751          \\ \hline
64 components   & 0.707   &   0.669         & 0.719          \\ \hline
           \\ 
%\hline
\end{tabular} \vspace{0.5em}
\end{table}

\subsection{street network}
\subsubsection{Visualisation experiments}

\begin{figure}[!ht]
	\centering 
	\includegraphics[width=1\linewidth] {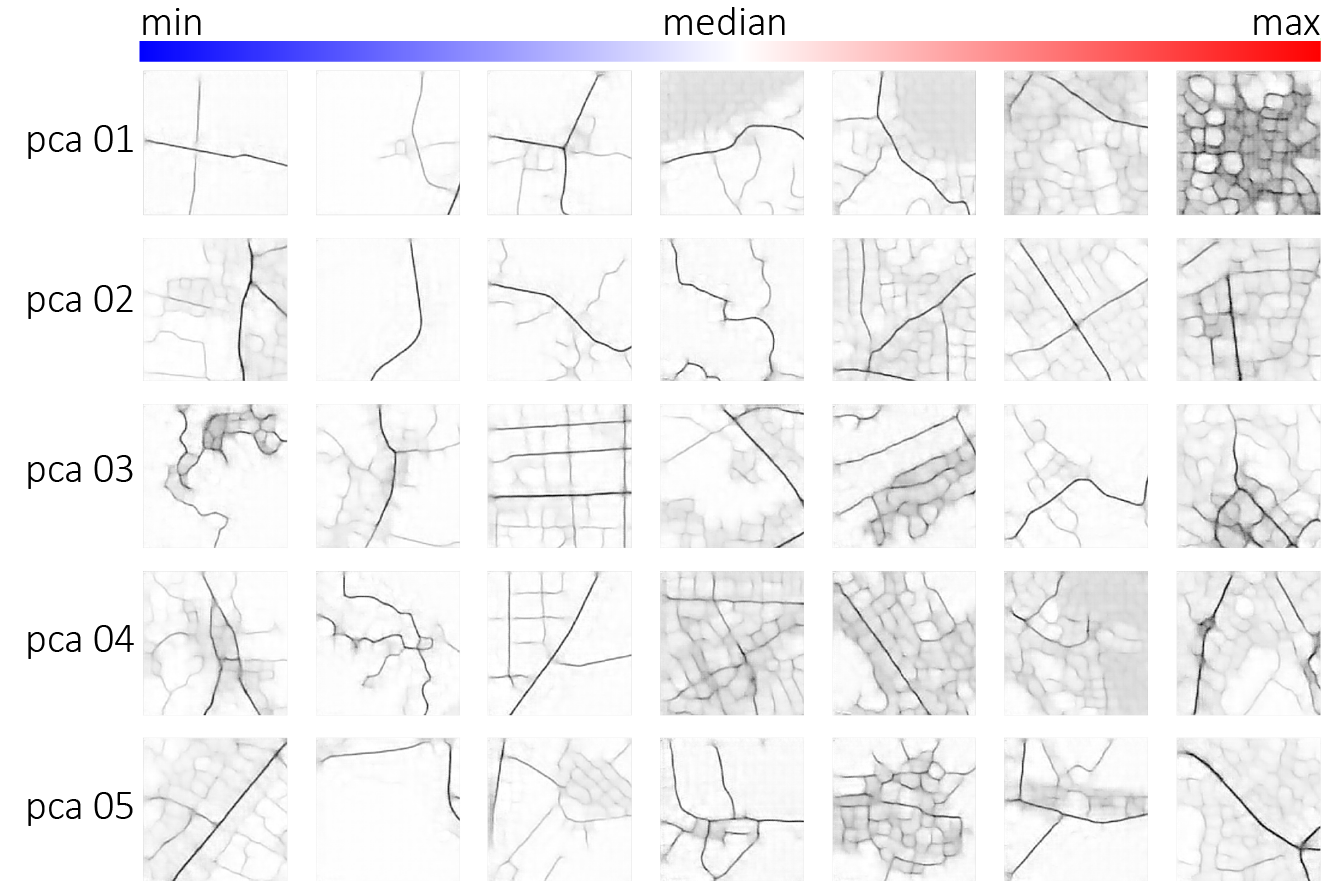}
	\caption{Example street network images for the first five principal components.}
	\label{fig:street_pca}
\end{figure}

For the street network case study, the trained convolutional autoencoder learned a mapping from the space of street network images (256 x 256 x 1 or 65,536 dimensions) to a lower dimensional latent space (640 dimensions) which are then further summarised into a set of linearly uncorrelated variables by applying ($PCA_{lin}$). By plotting out the street network images with the lowest to highest values of each component we can start to interpret the learnt latent space. In figure \ref{fig:street_pca}, we show the first five. These plots all relate to density of streets in different spatialised regions. The first $pca$ encodes general density, while $pca$ 2-5 encode spatialised densities (left-right, top-bottom, center-periphery, diagonals) respectively. 

\begin{figure}[!ht]
	\centering 
	\includegraphics[width=1\linewidth] {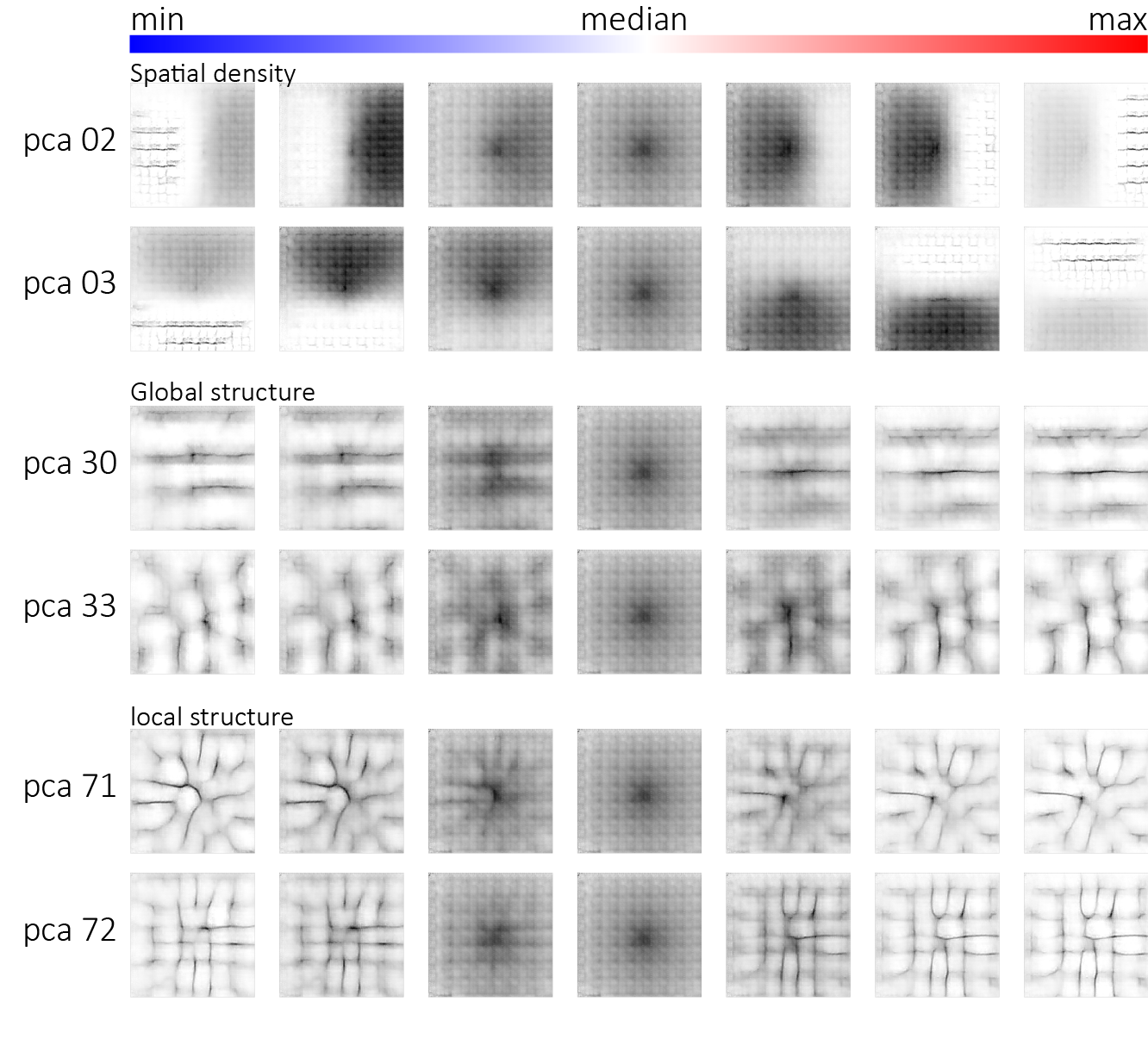}
	\caption{Latent visual component perturbation of an average street network image. By perturbing the visual component of an average image, we are able to show meaning of the perturbed component. The first sets of components seems to be related to spatialised density. While the second and third sets of components seems to be related to global and local structure of the street network.}
	\label{fig:streets_pca_median}
\end{figure}

To make it easier to interpret each $pca$ we create a mean vector $\hat{v}$, where we keep all values in $\hat{v}$ constant and vary only the $pca$ before passing it to the decoder to create a synthetic image. In figure \ref{fig:streets_pca_median}, we show a subset of the different latent visual components encoded by the $pca$ values. We show that the first 10 $pca$ encode regions of spatialised density. We confirm the clustered spatial structure of these component through a spatial autocorrelation tests. The results of the test are shown in the $appendix$ for the first 8 pca perturbations. $pca$ 11-50 encode global structure of the network (coarse grain detail), while $pca$ 50-640 encode local structure of the network (finer grain detail).

\begin{figure}[!ht]
	\centering 
	\includegraphics[width=1\linewidth] {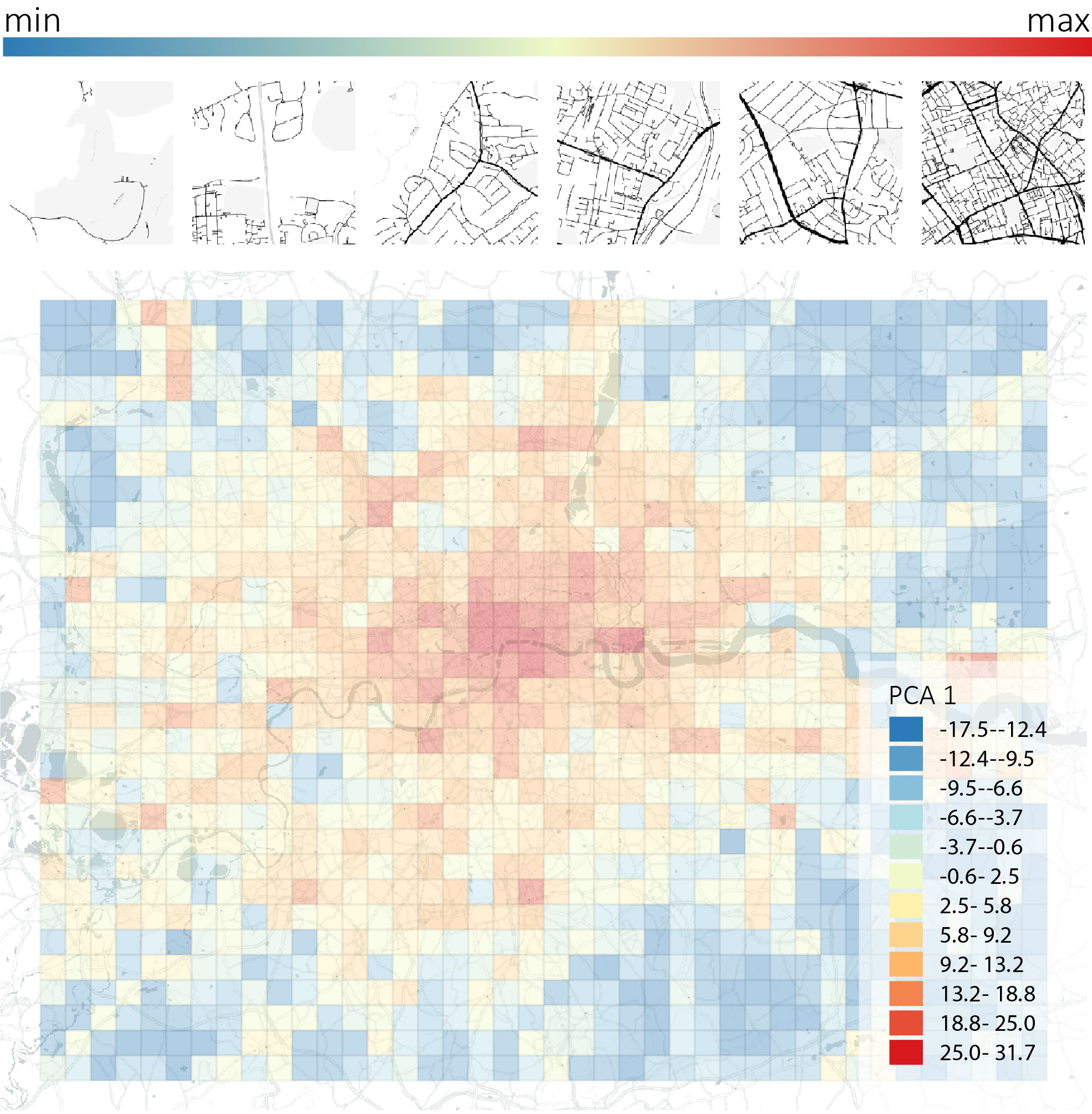}
	\caption{Values of the first principal component across London. The first component of the street network images can be interpreted as a measure of street network density.}
	\label{fig:street_map}
\end{figure}

By mapping the values of the principal components we can further test spatial patterns that they might encode. With just the first principal component of the latent space we are able to differentiate street network densities across the city of London. Figure \ref{fig:street_map} shows central London has higher street density than outer London.

\subsubsection{Prediction experiment}

Lastly, we test the ability these encoding can capture network features by using them to predict two network statistics: intersection density and closeness centrality. To do so, we first select a number of cities from our dataset where we retrieve its street network graphs $G=(V,E)$. For each graph, we calculate the closeness centrality of its nodes $u \in V$ through:

$$C(u)=\frac{n-1}{\sum_{v \in V}^{n-1}d(v,u)}$$

where $d(v,u)$ is the shortest weighted path between $u$ and $v$ and $n$ is the total number of nodes in the graph. We then create a continuous $1.5X1.5km$ rectangular grid over each city graph. For each grid cell, we define intersection density as the number of nodes inside the cell divided by the surface area, and the closeness centrality as its median values within each cell.

\begin{table}[!ht]
\centering
\caption{Street intersection density results \label{tab:streets1}}

\begin{tabular}{r|c|c|c}
%\toprule
%\hline
\textbf{}  $accuracy$      & $PCA_{lin}$      & $AE_{lin}$ & $AE_{non}$     \\ \hline
4 components    & 76.59\%   &   62.90\%         & 73.56\%         \\ \hline
8 components    & 77.28\%   &   76.43\%         & 72.01\%          \\ \hline
16 components   & 75.54\%   &   75.94\%         & 74.00\%         \\ \hline
32 components   & 71.15\%   &   71.65\%         & 76.00\%          \\ \hline
64 components   & 69.45\%   &   73.70\%         & 71.83\%          \\ \hline
           \\ 
%\hline
\end{tabular} \vspace{0.5em}

\begin{tabular}{r|c|c|c}
%\toprule
%\hline
\textbf{}  $losses$    & $PCA_{lin}$    & $AE_{lin}$ & $AE_{non}$     \\ \hline
4 components    & 0.23   &   0.37        & 0.25        \\ \hline
8 components    & 0.22   &   0.23        & 0.27          \\ \hline
16 components   & 0.26   &   0.23         & 0.26        \\ \hline
32 components   & 0.31   &   0.26        & 0.26          \\ \hline
64 components   & 0.35   &   0.24         & 0.25          \\ \hline
           \\ 
%\hline
\end{tabular} \vspace{0.5em}
\end{table}

\begin{table}[!ht]
\centering
\caption{Street closeness centrality results \label{tab:streets2}}

\begin{tabular}{r|c|c|c}
%\toprule
%\hline
\textbf{}  $accuracy$      & $PCA_{lin}$      & $AE_{lin}$ & $AE_{non}$     \\ \hline
4 components    & 54.63\%   &   59.89\%         & 60.23\%         \\ \hline
8 components    & 55.41\%   &   59.51\%         & 58.22\%          \\ \hline
16 components   & 54.17\%   &   58.19\%         & 59.02\%         \\ \hline
32 components   & 51.85\%   &   57.74\%         & 58.92\%          \\ \hline
64 components   & 34.33\%   &   52.00\%         & 53.45\%          \\ \hline
           \\ 
%\hline
\end{tabular} \vspace{0.5em}

\begin{tabular}{r|c|c|c}
%\toprule
%\hline
\textbf{}  $losses$    & $PCA_{lin}$    & $AE_{lin}$ & $AE_{non}$     \\ \hline
4 components    & 0.46   &   0.43        & 0.43        \\ \hline
8 components    & 0.45   &   0.41        & 0.45          \\ \hline
16 components   & 0.48   &   0.46         & 0.43        \\ \hline
32 components   & 0.53   &   0.47        & 0.41          \\ \hline
64 components   & 0.61   &   0.48         & 0.47          \\ \hline
           \\ 
%\hline
\end{tabular} \vspace{0.5em}
\end{table}

The data is then split into a train (70\%), validation (15\%), and test (15\%) set. We train a $mlp$ $F(\cdot)$ to predict both the intersection density and median closeness centrality for each grid cell for all our street network graphs from the visual latent components $v$. The $mlp$ here is defined by a fully connected neural network with two hidden layers, and a dropout layer (0.2) before the final activation. We define five different models based on the number of components [4,8,16,32,64] using the three dimension reduction techniques described in the methods section. 

The results in Table \ref{tab:streets1} show the $mse$ and $R^2$ for street intersection density using different number of $pca$ components. With just a few components we are able to achieve an accuracy of $77\%$ with a $PCA_{lin}$ model for the spatially random test-set on spatial features of the graph (intersection density), achieving slightly better results than both $AE_{lin}$ and $AE_{non}$. 

In the case of the median closeness centrality, shown in Table \ref{tab:streets2}, we achieve a $R^{2}$ of $55\%$ with a $PCA_{lin}$, showing we can achieve similar levels of accuracy to the other dimensionality reduction techniques. The difference in results between the intersection density and the median closeness centrality predictions is most likely due to the fact the while the intersection density is a local attribute thus can be entirely captured through the local graph structure within the grid cell, closeness centrality is dependent on the global structure of the graph of the entire city. Despite this, the model is informative and is still able to capture a significant portion of the variance of the closeness value with few components of the local graph structure. Further research is needed to investigate how much of the global structure can be inferred by local attributes.

% \begin{figure}
% 	\centering 
% 	\includegraphics[width=1\linewidth] {figs/streets_perturbations.jpg}
% 	\caption{PCA value perturbation}
% 	\label{fig:street_perturbation}
% \end{figure}

\section{Discussion and Conclusion}
We have presented a simple but novel unsupervised approach to extract and interrogate visual latent components from urban images. This exploratory research sits in contrast to previous works which focused on supervised learning \cite{Naik2013, Law2018, Gebru2017, Seresinhe2017}. Through geographic mapping, generative visualisations, and prediction experiments we were able to retrieve initial meaning from these visual latent components. With the increasing availability of large scale unlabelled image data, research into learning a compact representation automatically from geographical data will become increasingly useful.

In the case of the street level images, by mapping the visual latent components and generating synthetic images by perturbing its components, we were able to discover descriptive meaning from the data of which two of the primary components could be measures in describing street urbanity. We also found the lower dimensional latent components are able to predict two different generic urban characteristics such as street enclosure and street frontage type. The predictive accuracy for street frontage type is not as high as those using a purely supervised approach \cite{Law2018} but the results suggest a useful and generalisable representation can be learnt for different tasks. Despite the positive results, further exploration is necessary. For example, research is needed to relate the principal components to humanly labelled data describing the perception of street quality \cite{Naik2013}. The results can validate the meaning of these components quantitatively. To confirm the usefulness of the representation learnt, further research is also needed in comparing the visual latent components from unsupervised visual features with the visual latent components from supervised visual features (ie. Places365 database \cite{Zhou2014_places365}). Future researches are also needed on a) creating more realistic reconstructions by using generative models such as $VAE$ or $GAN$ b) developing quantitative methods to systematically disentangle and control interpretable latent components and c) conducting future research and designing experiments on semi-supervised learning and multi-task learning tasks.

In the case of the street networks, although the model is able to predict road network density and median closeness centrality, it fails to capture more complex street network features, we believe this is because the self-organized pattern of street networks is the result of both geometrical order/disorder as well as local rules of optimality. Through rasterising the street networks, the explicit topological data of the graph is lost, and the model is not able to recover this quality from the image alone. Future works can explore ways to incorporate topological properties of the networks into the model. Recent advances in graph neural networks provide promising directions that would allow both topological and geometric properties to be incorporated into the model, this would allow a richer representation of the street network as both local connectivity structure and their spatial embedding could be preserved. Despite its many limitations, there are benefits to such as approach where traditional network measures can sometimes be computationally expensive, for example $betweenness centrality$ has a time complexity of  $O(nm + n^{2}logn)$ and many spectral properties require eigenvalue decomposition of the graph laplacian matrix to be computed, with a time complexity of $O(n^3)$. A model that could approximate these parameters in an efficient manner could prove useful for varied applications, such as characterizing street networks across the world.

An immediate implication of the study, is that by learning a useful and compact representation from urban images, we can use this information immediately for other down-stream geographical tasks such as in prediction and classification. Conversely, this can reduce compute time and data collection costs significantly. More importantly though, the exploratory knowledge discovery process of using a combination of visualisation and inference, can shed new information about these non-linear methods such as neural networks and higher dimensional complex datasets such as images. To conclude, this research contributes to recent efforts in linking the disciplines of geography and machine learning. On the one hand, we find meaning from the visual latent components of street level and street network images. On the other hand, we also demonstrate how geographical datasets and visualisation techniques can be useful to enrich our understanding of machine learning methods. 

%in showing how machine learning methods can bring greater benefits to geographical research and reciprocally how geographical knowledge can bring greater meaning to machine learning methods.

%\section{Appendices}

%
% The acknowledgments section is defined using the "acks" environment (and NOT an unnumbered section). This ensures
% the proper identification of the section in the article metadata, and the consistent spelling of the heading.
%\begin{acks}
%To Robert, for the bagels and %explaining CMYK and color spaces.
%\end{acks}

%
% The next two lines define the bibliography style to be used, and the bibliography file.
\bibliographystyle{ACM-Reference-Format}
\bibliography{sigconf}

%\pagebreak

\section{Appendix}
\label{appendix}
In the appendix, we describe for both the StreetView case study and the Street Network case study, the architecture of the Convolutional Autoencoder, the stacked autoencoders, the multi-layer-perceptron ($mlp$) and the spatial autocorrelation analysis. %which maps the image $i$ to its reconstruction $i^*$ via a latent variable $z$, the two stacked autoencoder which reduces the dimension of $z$ down to $v$ and the multi-layer-perceptron which maps the latent component $v$ to both the street-level and street-network attributes. We then describe some additional spatial analysis we have conducted on the latent space, in this case studying the spatial autocorrelation structure. 

\subsection{StreetView architecture}
\subsubsection{Convolutional Auotoencoder architecture}
For the Convolutional Autoencoder of StreetView, the input is a fixed sized $224 x 224$ three channel coloured image. We adopted a simplified convolution blocks from $VGG$ \cite{VGGnet} as the basis of the architecture where we keep the kernel size and filter numbers constant across both the encoder and decoder. Let $Ck$ denote a Convolution-ReLU layer with $k$ filters and $Cdk$ denotes a Convolutional-ReLu-Upsample layer with $k$ filters. All convolutions are $3×3$ spatial filters applied with stride of 1. 

encoder:C64-C64-C128-C128-C256-C256-C512-C512-CC512-C512-C512-CC512

decoder:CD512-CD512-CD512-C512-C512-C512-C256-C256-C256-C128-C128-C64-C64

%decoder:C3-C32-C64-C128-C256-256
%decoder:Cd256-Cd256-Cd128-Cd64-Cd32-Cd3

\subsubsection{Stacked Autoencoder architecture}
For the stacked Autoencoder, where we summarise the latent variable $z$ to its latent component $v$, we applied two forms of the autoencoder namely a linear autoencoder and a nonlinear autoencoder. The linear autoencoder uses linear activation functions with one bottleneck layer that outputs $V$. The nonlinear autoencoder on the otherhand uses the $ReLu$ activation function with three hidden layers where the first and the third hidden layer are the encoding and decoding layer with $512$ neurons and the second layer being the bottleneck layer that outputs $V$. Let $Dk$ denote a Dense-ReLU layer with $k$ filters and $N$ as the number of components in the bottleneck layer.

linear: 4096-N-4096

non-linear: 4096-D512-N-D512-4096

\subsubsection{Multi-layer Perceptron}

%decoder:CD3-C32-C64-C128-C256-256
%decoder:CD256-C256-C128-C64-C32-C3

The multi-layer perceptron $mlp$ here is defined as a fully connected neural network with three hidden layers. The first fully connected layer has 64 hidden nodes, the second has 32 hidden nodes, while the third layer has 16 hidden nodes. A dropout layer ($0.2$) and $l1$ regularisation was added in the final activation layer. We constructed five different models based on the number of components [4,8,16,32,64] and based on the three dimension reduction techniques resulting in a total of 15 models. Let $Dk$ denote a Dense-ReLu layer with $k$ number of neurons and $N$ denote the shape of the visual latent component. [4,8,16,32,64].

Multi-layer perceptron: N-D64-D32-D16-1

\subsection{Street network architecture}
\subsubsection{Convolutional Auotoencoder architecture}
For the Convolutional Autoencoder of street network data, the input is a fixed sized $256 x 256$ single channel gray-scale.  We use a stack of convolutional-ReLu layers and transposed convolutional layers, with a fixed small receptive field: $3X3$ and a convolution stride fixed to 2 pixels. Let $Ck$ denote a Convolution-ReLU layer with $k$ filters and $TCk$ denotes a Transposed-Convolution-ReLu layer with $k$ filters.

encoder:C15-C15-C15-C10-C10

decoder:TC10-TC10-TC15-TC15-TC1

\subsubsection{Stacked Autoencoder architecture}
For the stacked Autoencoder, where we summarise the latent variable $z$ to its latent component $v$, we applied two forms of the autoencoder namely a linear autoencoder and a nonlinear autoencoder. The linear autoencoder uses linear activation functions with one bottleneck layer that outputs $V$. The nonlinear autoencoder on the otherhand uses the $ReLu$ activation function with three hidden layers where the first and the third hidden layer are the encoding and decoding layer with $128$ neurons and the second layer being the bottleneck layer that outputs $V$. Let $Dk$ denote a Dense-ReLU layer with $k$ filters and $N$ as the number of components in the bottleneck layer.

linear: 640-N-640

non-linear: 640-D128-N-D128-640

\subsubsection{Multi-layer Perceptron}

The multi-layer perceptron $mlp$ here is defined as a fully connected neural network with two hidden layers. The first fully connected layer has 32 hidden nodes, while the second layer has 16 hidden nodes. A dropout layer ($0.2$) was added before the final activation layer and a $l1$ regularisation was added in the final activation layer. We constructed five different models based on the number of components [4,8,16,32,64] and based on the three dimension reduction techniques. Let $Dk$ denote a Dense-ReLu layer with $k$ number of neurons and $N$ denote the shape of the visual latent component. [4,8,16,32,64].

Multi-layer perceptron: N-D32-D16-1

\subsection{Global Spatial Autocorrelation structure}

\subsubsection{Street network images}
In the case of the rasterized street network data, we test if the latent components capture strong local spatial inter-dependencies. This can be examined by measuring the autocorrelation of pixels with its local neighbours when perturbing the principal components of an average image. For our purposes we assume that the output of our $ConvPCA$ $I'$ follow some spatial process $y \sim f(c)$, where $y=vec(I')$ and $c$ is a vector indexing the $y_{i}$ pixel values in the output $I'$.  The local spatial autocorrelation $L_{i}= L(y_{i})$ is computed as:

$L_{i}=(n-1) \frac{y_{i}-\bar{y}}{\sum_{j=1, j\neq i}(y_{j}-\bar{y})^{2}} \sum_{j=1, j \neq i}w_{i,j}(y_{j}-\bar{y})$

where $\bar{y}$ represents the mean of $y_{i}$'s and $w_{i,j}$ are components of a weight matrix indicating membership of the local neighbourhood set between pixels $i$ and $j$. Below, we show the results of the global spatial autocorrelation $L=\sum_{i}L_{i}$ of the max and minimum perturbations of the first 8 $pca$ values and their corresponding $I's$. 

\begin{table}[H]
\centering
\caption{Global spatial autocorrelation of outputs of min-max perturbations of first 8 PCA's of street networks \label{tab:moran's I}}

\begin{tabular}{r|c|c}
%\toprule
%\hline
\textbf{}  $PCA$    & $I of min$  & $I of max$  \\ \hline
$1^{st}$ component    & 0.87   &   0.88 \\ \hline
$2^{nd}$ component    & 0.96   &   0.87 \\ \hline
$3^{rd}$ component   & 0.93   &   0.89 \\ \hline
$4^{th}$ component   & 0.98   &   0.94 \\ \hline
$5^{th}$ component   & 0.99   &   0.97 \\ \hline
$6^{th}$ component   & 0.98   &   0.96 \\ \hline
$7^{th}$ component   & 0.99   &   0.98 \\ \hline
$8^{th}$ component   & 0.99   &   0.98 \\ \hline
%           \\ 
%\hline
\end{tabular} \vspace{0.5em}
\end{table}

\subsubsection{StreetView images}
In the case of the street level images, we test if the latent components exhibit strong geographical dependencies. This can be examined by measuring the spatial autocorrelation between a street component value with its local neighbours, in this case defined by its $8^{th}$ nearest local neighbours. The global Moran's I $L$ for the two primary components are calculated. The result shows a strong spatial dependence of the visual latent component values at the street level.

\begin{table}[H]
\centering
\caption{Global spatial autocorrelation of $pca$ components \label{tab:moran's I}}

\begin{tabular}{r|c}
%\toprule
%\hline
\textbf{}  $PCA$    & $L$    \\ \hline
$1^{st}$ component    & 0.68    \\ \hline
$3^{rd}$ component    & 0.75   \\ \hline
           \\ 
%\hline
\end{tabular} \vspace{0.5em}
\end{table}

\end{document}